\documentclass[runningheads]{llncs}
\usepackage[T1]{fontenc}
\usepackage{graphicx}
\usepackage{booktabs}
\usepackage[misc]{ifsym}
\newcommand{\corr}{(\Letter)}
\usepackage{mwe}
\usepackage{amsmath}
\usepackage{amsfonts}
\usepackage{subfigure}
\usepackage{color}
\usepackage{algorithm}
\usepackage{hyperref}
\usepackage{enumitem}
\usepackage{algpseudocode}
\usepackage{multirow}
\usepackage{array}

\begin{document}

\title{Colliding with Adversaries at ECML-PKDD 2025 Model Robustness  Competition 
 \\ 1st Prize Solution
}

\titlerunning{ Colliding with Adversaries  -  Model Robustness 1st Prize Solution}

\author{Dimitris Stefanopoulos\inst{1} \corr \and Andreas Voskou\inst{2} }

\institute{Aristotle University of Thessaloniki \email{dstefanop@math.auth.gr}  \and  Cyprus University of Technology, Limassol, Cyprus  \email{ai.voskou@edu.cut.ac.cy} } 

\maketitle 

\begin{abstract}
This report presents the winning solution for Task 2 of Colliding with Adversaries: A Challenge on Robust Learning in High Energy Physics Discovery at ECML-PKDD 2025. The goal of the challenge was to design and train a robust ANN-based model capable of achieving high accuracy in a binary classification task on both clean and adversarial data generated with the Random Distribution Shuffle Attack (RDSA). Our solution consists of two components: a data generation phase and a robust model training phase. In the first phase, we produced 15 million artificial training samples using a custom methodology derived from Random Distribution Shuffle Attack (RDSA). In the second phase, we introduced a robust architecture comprising (i)a Feature Embedding Block with shared weights among features of the same type and (ii)a Dense Fusion Tail responsible for the final prediction. Training this architecture on our adversarial dataset achieved a mixed accuracy score of 80\%, exceeding the second-place solution by two percentage points.
\end{abstract}

\section{Introduction}

The task focuses on developing a deep neural network that is robust to adversarial examples in high-energy physics data presented in tabular format. Specifically, it addresses a binary classification problem that aims to distinguish between two classes of particle jets produced by particle colliders such as the LHC (Large Hadron Collider): "two top-jets (TTJets)" and "two W-boson-jets".

Given that the RDSA (Random Distribution Shuffle Attack) method is used to generate adversarial examples for TopoDNN (\cite{kasieczka2019machine},\cite{saala2025introduction}), we opted for a method-aware robustness strategy. In particular, we applied an antiRDSA approach that perturbs data points using the same technique employed to generate the adversarial evaluation set, while preserving their original labels. As a result, our augmented training set includes multiple correctly labeled, perturbed adversarial examples. 

The final robust model is an ensemble of four classifiers combined with averaging. Each classifier incorporates distinct feature embedding layers for the three physically different variables: transverse momentum ($p_T$), azimuthal angle ($\phi$), and pseudorapidity ($\eta$). The classifiers differ in their input dropout rates and training datasets, which correspond to two separate augmentation phases. The full code for data augmentation, model architecture and training can be found in the \href{https://github.com/jmstf94/Colliding-with-Adversaries}{Colliding with Adversaries GitHub repository}.

\begin{algorithm}[h]
\caption{antiRDSA Data Generation from Feature Distributions}
\begin{algorithmic}
\State \textbf{Input:} Dataset $D$ with $87$ features
\State \textbf{Output:} List of generated variants $V$
\State $GenData \gets [\ ]$
\State Compute histograms for each feature in $D$ with $n_{\text{bins}}$ bins
\For{each selected sample $x$ in $D$}
    \For{$i = 1$ to $50$}
        \State $x' \gets x$
        \State Randomly select $n_{\text{vars}}$ feature indices $\{f_1, \ldots, f_{n_{\text{vars}}}\}$
        \For{each feature index $f$ in $\{f_1, \ldots, f_{n_{\text{vars}}}\}$}
            \State Sample value $v_f$ from empirical distribution of feature $f$
            \State $x'[f] \gets v_f$
        \EndFor
        \State Append $(x', \text{label}(x))$ to $V$
    \EndFor
\EndFor
\end{algorithmic}
\end{algorithm}

\section{Problem Statement}
Given a dataset $D$ with finite number of $|D| = n$ data points  $d \in \mathbb{R}^{87} \times \{0,1\}$, our objective is to generate a larger dataset of size $N \gg n$ in order to train a robust model $F: \mathbb{R}^{87} \to \{0,1\}$ that minimizes the following metric:
\begin{equation}
S = \frac{\textit{Clean Accuracy} + \textit{Adversarial Accuracy}}{2}
\end{equation}
Here, \textit{Clean Accuracy} and \textit{Adversarial Accuracy} represent the proportion of correctly classified instances (including both true positives and true negatives) in two separate hidden evaluation sets. Although examples of Clean and Adversarial test sets were provided, improvements on these sets did not fully transfer to the corresponding hidden evaluation sets on the platform. In particular, Adversarial Accuracy showed a significant decrease, likely by design, to assess the generalization capabilities of the robust methodology.
\section{Data Generation}

The data generation methodology is largely based on the Random Distribution Shuffle Attack (RDSA) \cite{saala2025enforcing}, which was employed by the competition committee to create adversarial examples using a TopoDNN model. As the original algorithm relies on TopoDNN predictions to generate adversaries, our modified version of RDSA (so called \textit{antiRDSA}) excluded the model query step and produced several large data points to assist in training. This data points contain a significant number of perturbed inputs, each labeled with the correct class. Consequently, the resulting data consist of many examples that resemble adversarial instances but are annotated with the true label. 

The final augmented training sets were generated using data from the \href{https://huggingface.co/datasets/TSaala/CollidingAdversaries/tree/main}{HuggingFace repository TSaala/CollidingAdversaries}. The generation algorithm was applied to the Train, Val, and Test splits twice with different parameter settings. \autoref{datagen} presents the parameters used for generating DataGen1 and DataGen2. The column \textit{Source} indicates the HuggingFace dataset split used as the basis for each generation phase. The number of variants per sample is fixed at 50, and the subset size always matches the full length of the respective "Source" dataset.
\begin{table}[h!]
\caption{Description statistics for the two data generation cases.}
\centering
\begin{tabular}{ccccc}
\toprule
\textbf{Name$\;$} & \textbf{Source$\;$} & \textbf{nBins$\;$} & \textbf{nVars$\;$} & \textbf{Size($\times 10^6$)} \\
\midrule
\multirow{3}{*}{DataGen1} & Train & 100 & 5 & 5 \\
                         & Val & 200 & 5 & 1.25 \\
                         & Test & 200 & 5 & 1.25 \\
\midrule
\multirow{3}{*}{DataGen2} & Train & 100 & 10 & 5 \\
                         & Val & 200 & 10 & 1.25 \\
                         & Test & 200 & 10 & 1.25 \\
\bottomrule
\end{tabular}
\label{datagen}
\end{table}
In \autoref{dataaug}, the final training datasets are presented. These are heavily augmented versions of the Hugging Face datasets. DataAug1 and DataAug2 correspond to the training sets used for the successful ensemble models.
\begin{table}[h!]
\caption{Overview of the final augmented datasets used for training. The "Total Size" column reflects the combined size of the original HuggingFace dataset and the synthetic data generated as described in \autoref{datagen}.}
\centering
\begin{tabular}{cccc}
\toprule
\textbf{Name} & \textbf{HuggingFace data} & \textbf{Generated Data} & \textbf{Size($\times 10^6$)} \\
\midrule
DataAug1 & Train+Val+Test  & DataGen1 & 7.65 \\
DataAug2 & Train+Val+Test  & DataGen1+DataGen2 & 15.15 \\
\bottomrule
\end{tabular}

\label{dataaug}
\end{table}

\section{Robust Model}
\label{sec4}

\begin{figure}[h] 
    \centering
    \includegraphics[width=0.7\textwidth]{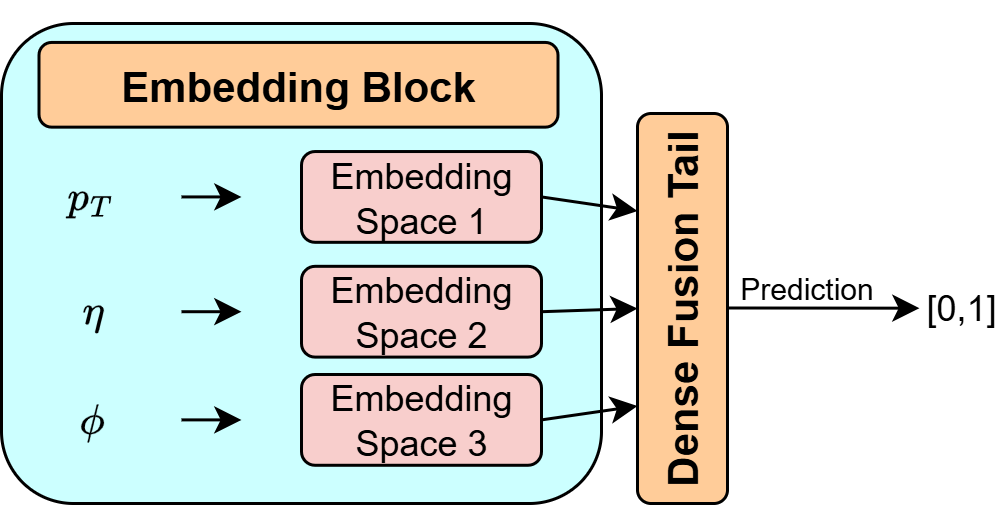} 
    \caption{Robust model architecture schema.}
    \label{fig:example}
\end{figure}

The architecture adopted for this robust model follows the structure of feature embeddings combined with a simple MLP (Multi-Layer Perceptron) for classification, consistent with the approach proposed in \cite{gorishniy2022embeddings}. The feature embedding component performs a non-linear transformation of each of the 87 scalar input features into a representation vector. This process can be formally described using $f_e : \mathbb{R} \rightarrow \mathbb{R}^8$. An abstract picture of the overall architecture is shown in \autoref{fig:example}.

While state-of-the-art methods employ periodic

, bin-based, or stochastic embedding mechanisms \cite{gorishniy2022embeddings,voskou2024transformers,gorishniy2021revisiting,rahat2024volvo}, such approaches could not be applied due to constraints imposed by the competition format. Instead, we implemented a simpler two-layer ReLU-activated embedding, using separate weights for each of the three input feature types: $p_T$, $\phi$, and $\eta$. The embedding of each of the 87 features is defined as:
\begin{equation}
\boldsymbol{x_e} = f_e(x \mid W_{\text{1}}^{t}, W_{\text{2}}^{t}) = W_{\text{2}}^{t} \text{ReLU}(W_{\text{1}}^{t}  x)
\end{equation}
where  $x \in R$, $\boldsymbol{x_e} \in R^8$, $W_{\text{1}}^{t} \in \mathbb{R}^{16 }$, $W_{\text{2}}^{t} \in \mathbb{R}^{8 \times 16}$, and $t \in \{\eta, \phi, p_T\}$.

In the Dense Fusion Tail, the representation vectors of all features are concatenated into a single vector of dimension $D = 87 \cdot 8$, and passed to an MLP classifier with one hidden layer of size 256, using the $\tanh$ activation function. The full model reads as:
\begin{equation}
      F(\boldsymbol{x}) =   W_{out}tanh( W_{hidden}(f_e(x_i))_{i=1}^{87} )
\end{equation}
where $W_{hidden}$ and $W_{out}$ the trainable parameters of the first and second layer of the MLP classifier. 

Dropout with $p=0.2$ and Gaussian noise with standard deviation $\sigma=0.01$ are applied to the hidden layers of both the MLP and the embedding network. Additional dropout is also applied to the raw input data.

The final model is an averaging ensemble of 2+2 such networks. One pair uses input dropout $p=0.075$, and the other uses $p=0.125$. Within each pair, one model is trained on both DataAug1 and DataAug2 while the other is trained only on  DataAug1. 

Each model was trained for a single epoch on its corresponding training dataset, without any form of validation. This decision was motivated by the absence of a reliable method for validating models on adversarial data—the hidden test set appeared to differ significantly from any dataset we were provided or could generate ourselves. Additionally, we utilized all the available Train, Validation, and Test data, in both their clean and augmented forms, for training purposes. Given the resulting size of the dataset, training for one epoch was sufficient.

\section{Discussion }

The core rationale behind our approach is that access to adversarial data with properties matching, or closely approximating, those of the test set is essential for training a robust model. Because no samples of similarly generated data were provided, we implemented a scalable approximation of the adversarial generator described in the available materials, enabling efficient synthesis of large volumes of training data.

Our model’s architecture builds on recent work showing the value of feature embedding for tabular data. We adapt the PLR-MLP \cite{gorishniy2022embeddings} formulation, substituting its periodic PLR embedding layer with a ReLU-activated alternative due to format constraints in the competition. Rather than fully sharing embedding parameters across all features or assigning completely separate parameters to each, we take a middle path: we group input features by their three distinct data types and learn one shared set of embedding parameters per type. This design reduces model size and inference cost while retaining more representational flexibility than a single shared embedding, striking a balance between parameter efficiency and expressiveness.

For future work, we believe performance could be improved by approximately 0.02–0.03 units through the integration of more modern architectures, such as the original periodic embedding variants, transformer-based models \cite{voskou2024transformers,gorishniy2021revisiting}, and other advances in tabular data modeling. Those examples have demonstrated much better results than simple MLPs and similar models on numerous standard benchmarks of tabular datasets, including physics-related material like the Higgs and Higgs-small datasets. Additionally, exploring stochastic methodologies, including Bayesian variational deep learning techniques \cite{panousis2021wta}, may further enhance performance by increasing the model’s robustness to adversarial or distributional shifts.

\section{Conclusion}
Using the antiRDSA approach for data generation and employing neural architectures utilizing different feature embeddings for each of the 3 feature types ($p_T$, $\phi$ or  $\eta$), we achieved the highest score of $S=0.8$, securing first place in the competition with a margin of $0.02$ over the second-best solution. The accuracy on clean samples was $0.88$, while the \textit{Adversarial Accuracy} score was approximately $0.72$, averaging to to the winning score.

While our robust methodology achieved competitive scores, we believe that extending the accepted model formats to support modern model architectures and arbitrary custom layers could further enhance performance and robustness against adversaries.

\bibliographystyle{splncs04}
\bibliography{paper}

\end{document}